\documentclass[10pt,twocolumn,letterpaper]{article}
\usepackage{wacv}
\usepackage{times}
\usepackage{epsfig}
\usepackage{graphicx}
\usepackage{amsmath}
\usepackage{amssymb}
\usepackage{booktabs}
\usepackage{float}
\usepackage{pifont}% http://ctan.org/pkg/pifont
\usepackage{multirow}
\usepackage[accsupp]{axessibility}  % Improves PDF readability for those with disabilities.
\newcommand{\xmark}{\ding{55}}%

\usepackage{xcolor}

\usepackage{caption}
\usepackage{subcaption}

\def\wacvPaperID{1952} % Enter the WACV Paper ID here

\wacvalgorithmstrack   % Uncomment this line if you are submitting to the Algorithms Track.
\wacvfinalcopy % *** Uncomment this line for the final submission

\usepackage{dsfont}
\usepackage{etoolbox}

\newif\ifshowedits

\newcommand{\addeditor}[3]{%
  \definecolor{#1color}{rgb}{#3}
  \expandafter\newcommand\csname #1\endcsname[1]{%
  \ifshowedits
    {\color{#1color} ##1}%
  \else
    {##1}%
  \fi
  }%
  \expandafter\newcommand\csname #1rmk\endcsname[1]{%
  \ifshowedits
    {\color{#1color} {\bf [#2: ##1]}}
  \fi
  }%
  \expandafter\newcommand\csname #1rpl\endcsname[2]{%
  \ifshowedits
    {\color{#1color} ##1 \sout{##2}}
  \else
    {##1}
  \fi
  }%
}

\newcommand{\createtextvar}[1]{
  \expandafter\newcommand\csname #1\endcsname{%
  {\text{#1}}
}%
}
\newcommand{\textvars}[1]{\forcsvlist{\createtextvar}{#1}}

\newcommand{\moretextwithfigures}{
\renewcommand{\topfraction}{1}
\renewcommand{\dbltopfraction}{1}
\renewcommand{\bottomfraction}{1}
\renewcommand{\textfraction}{.0}
\renewcommand{\floatpagefraction}{1}
\renewcommand{\dblfloatpagefraction}{1}
}

\newcommand{\mycomment}[1]{}

\newcommand{\calL}{{\cal L}}

\showeditsfalse

\moretextwithfigures

\ifwacvfinal
\usepackage[breaklinks=true,bookmarks=false]{hyperref}
\else
\usepackage[pagebackref=true,breaklinks=true,colorlinks,bookmarks=false]{hyperref}
\fi

\begin{document}

%%%%%%%%% TITLE

\title{Automatically Annotating Indoor Images with CAD Models via RGB-D Scans}

\author{Stefan Ainetter\textsuperscript{(1)}, Sinisa Stekovic\textsuperscript{(1)}, Friedrich Fraundorfer\textsuperscript{(1)}, Vincent Lepetit\textsuperscript{(2,1)}
 \\
\textsuperscript{(1)}Institute for Computer Graphics and Vision, Graz University of Technology, Graz, Austria\\
\textsuperscript{(2)}LIGM, \'Ecole des Ponts, Univ Gustave Eiffel, CNRS, Marne-la-Vallée, France\\
{\tt\small \{stefan.ainetter, sinisa.stekovic, fraundorfer\}@icg.tugraz.at}, {\tt\small vincent.lepetit@enpc.fr}
% For a paper whose authors are all at the same institution,
% omit the following lines up until the closing ``}''.
% Additional authors and addresses can be added with ``\and'',
% just like the second author.
% To save space, use either the email address or home page, not both
%\and
%Second Author\\
%Institution2\\
%First line of institution2 address\\
%{\tt\small secondauthor@i2.org}
}

\maketitle
\thispagestyle{empty}
%%%%%%%%% ABSTRACT

\begin{abstract}
We present an automatic method for annotating 
images of indoor scenes with the CAD models of the objects by relying on 
RGB-D scans.  Through a visual evaluation by 3D experts, we show that our method retrieves annotations that are at least as accurate as manual annotations, and can thus be used as ground truth without the burden of manually annotating 3D data.
We do this using an analysis-by-synthesis approach, which compares renderings of the CAD models with the captured scene. We introduce a 'cloning procedure' that identifies objects that have the same geometry, to annotate these objects with the same CAD models.
This allows us to obtain complete annotations for the ScanNet dataset and the recent ARKitScenes dataset. Source code and data will be available at \url{https://github.com/stefan-ainetter/SCANnotate}.

\end{abstract}

%%%%%%%%% BODY TEXT

\section{Introduction}
3D scene understanding is one of the most challenging problems in computer vision. For indoor scenes, several datasets are already available including SceneNN~\cite{hua2016scenenn}, ScanNet~\cite{dai2017scannet}, Matterport3D~\cite{Matterport3D}, and ARKitScenes~\cite{dehghan2021arkitscenes}. However, with the exception of ScanNet for which some of the objects are annotated thanks to the Scan2CAD dataset~\cite{avetisyan2019scan2cad},  they do not provide annotations for the shapes of the objects.  This is because 3D manual annotations of the shapes are particularly cumbersome to create, as one has to jointly estimate 
a good 6D pose and retrieve a well suitable CAD model for each object.
The Pix3D dataset~\cite{sun2018pix3d} is also annotated with CAD models. However, it is made of single-object images, without partial occlusions, which facilitates the annotations but is also likely to bias training and evaluation.

\begin{figure}[!]
 %trim={<left> <lower> <right> <upper>}
      \centering
  		\includegraphics[trim={0cm 0.cm 0cm 0.cm},width=.915\linewidth]{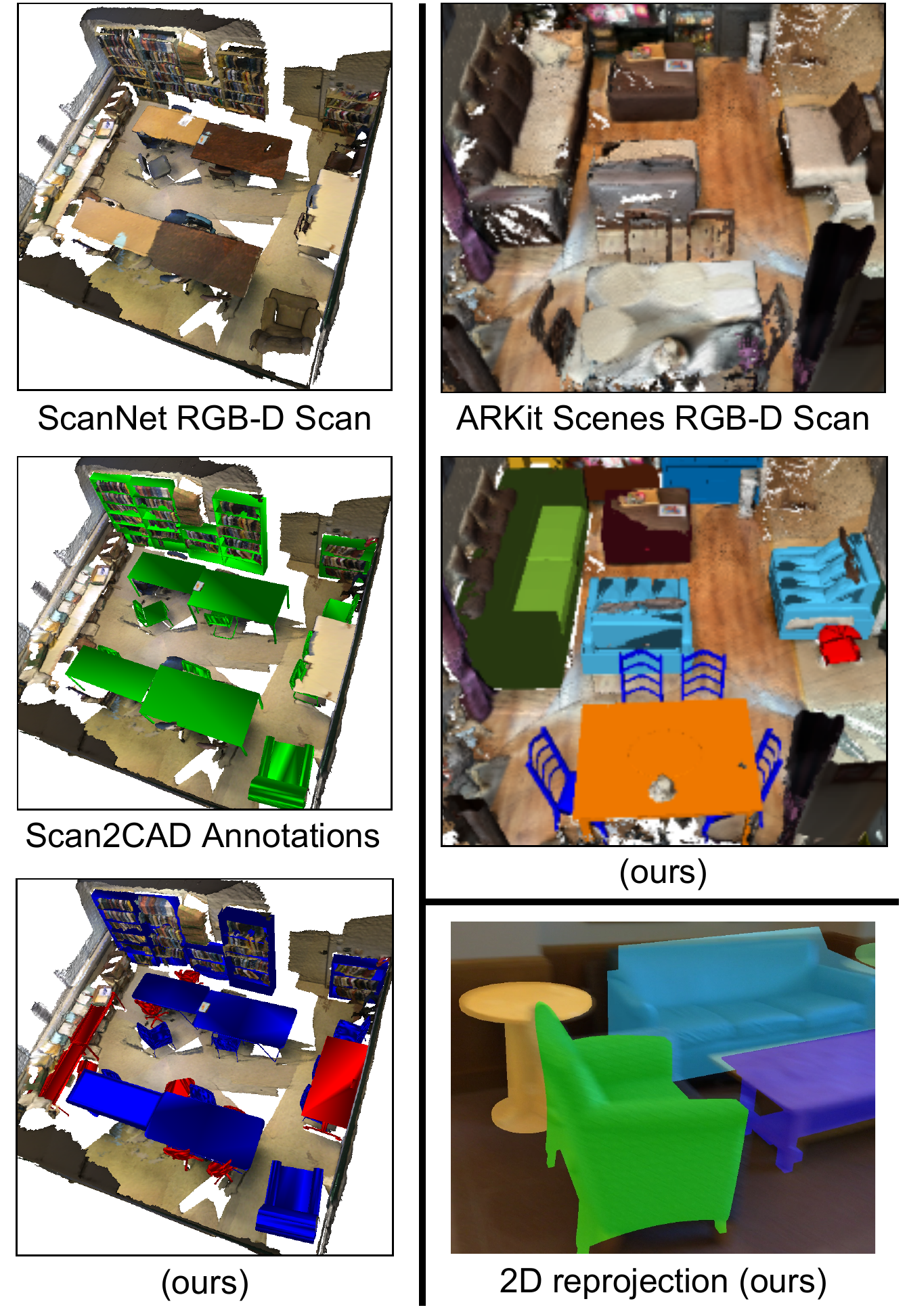}
	\caption{Given an RGB-D video sequence of a scene and 3D
          oriented bounding boxes or 3D semantic instance
          segmentation, our method retrieves CAD models from ShapeNet
          that fit the objects well. \textbf{Left: } Scan2CAD does not
          provide CAD models for all the objects; The CAD models in
          red are the models we retrieve for the objects not annotated
          by Scan2CAD.
          \textbf{Right: } Our cloning procedure retrieves the same
          CAD model for the objects sharing a common shape, such as
          the chairs in this example.
          \textbf{Bottom right: } The CAD models and the poses we
          retrieve reproject well in the images, and can therefore be
          used for training and testing fast inference methods, for
          example methods that predict the geometry of the objects from single images. 
	}
	\label{fig:teaser}
		\vspace{-24pt}
\end{figure}
Using synthetic images for training is another option, however generating realistic virtual 3D scenes also has a high financial cost, both in terms of creation and rendering~\cite{applepaper}. Moreover, testing should still be done on real images.

In this paper, we report our work on how to automatically retrieve good CAD models for objects in a scene captured by a moving RGB-D camera. Figure~\ref{fig:teaser} shows the CAD models we retrieved for one scene from ScanNet and one scene from ARKitScenes. Visual evaluation by experts in the domain of computer vision demonstrated that our CAD models are often as good or better than the manual annotations from Scan2CAD, while they are obtained automatically. We thus see our method as a tool that automatically generates annotations, which can then be used to train supervised methods for fast inference. On average, our non-optimized implementation takes approximately $11.00$ minutes per scene using two  NVidia  GeForce  RTX  2080  Ti graphics  cards, compared to Scan2CAD human annotations that take $20.52$ minutes per scene~\cite{avetisyan2019scan2cad}, while being completely automated.

Our method is not based on learning and thus does not require registered 3D models for training. Instead, we use as input 3D oriented bounding boxes or instance segmentation for the objects: These annotations are much simpler than CAD model fitting and they are in fact available for the main existing datasets as detailed in Table~\ref{tab:datasets_overview}. Because we also adjust the bounding boxes anyway, these boxes do not have to be particularly accurate. To find a suitable CAD model, we sequentially replace the target object with one of the 3D models from the ShapeNet database~\cite{chang2015shapenet}. By following an analysis-by-synthesis approach, we select the best CAD model by comparing depth renderings to the captured depth maps while adjusting the 9D pose and scale of the CAD models. As Figure~\ref{fig:teaser} also shows, by relying on depth data to select and register the CAD models, the CAD models we retrieve are not only well located in 3D but also reproject well in the images of the RGB-D scans. 

In man-made environments, it often happens that multiple objects have the same shape, such as the chairs around the table for the ARKitScenes example in Figure~\ref{fig:teaser}. However, independently searching for the CAD models of these multiple instances is likely to yield different 3D models. To exploit the high-level knowledge that several objects tend to have the same shape, we cluster the CAD models retrieved after an independent search based on the similarity of their shapes. We then perform a joint retrieval, looking for the same CAD model for all the objects in the cluster. This way, we can exploit more information from the depth maps to retrieve a better CAD model:  For example, if we found that two chairs share the same shape, but the top of the first chair and the bottom of the second chair are not visible in the RGB-D scan, we can still recover a correct CAD model for the two chairs. 

\begin{figure*}[t] 
 %trim={<left> <lower> <right> <upper>}
      \centering
 		\includegraphics[trim={0cm 0.cm 0cm 0.cm},width=0.9\linewidth]{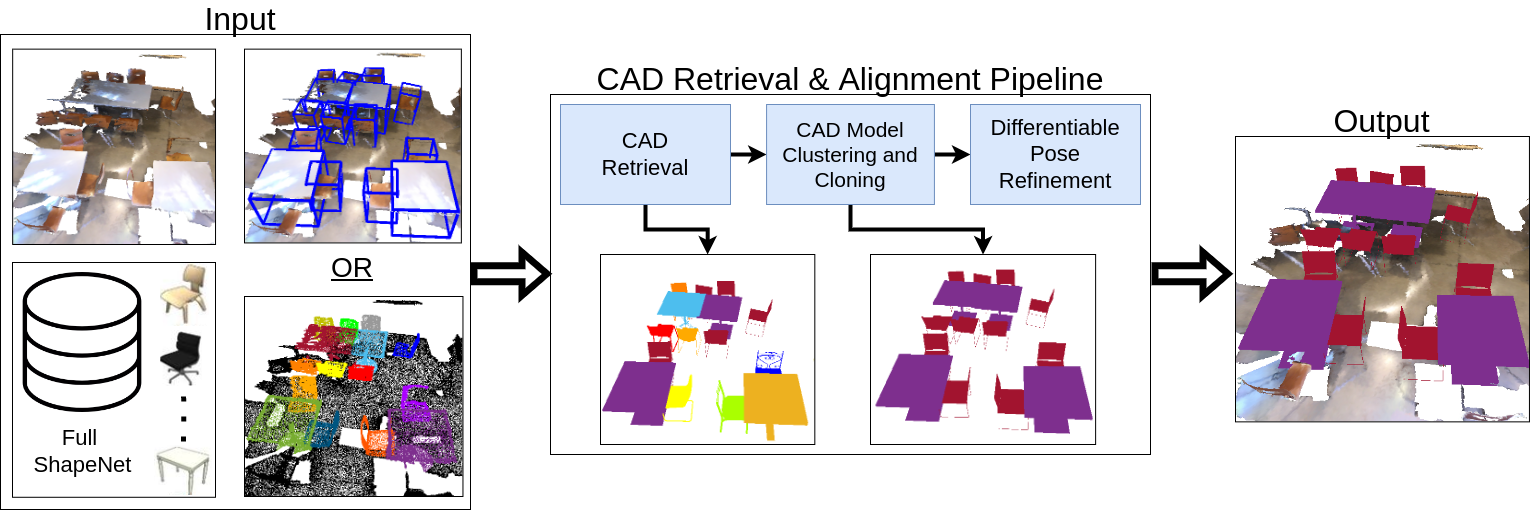}
	\caption[]{\textbf{Overview of our method.} We use as input the RGB-D scan of an indoor scene, and either 3D box annotations or 3D object instance segmentation to accurately retrieve and align CAD models from ShapeNet. We also retrieve the same CAD model for objects with the same shape: 3D models renderings with the same color correspond to the same CAD model.}
	\label{fig:overview}
\end{figure*}

We did not find any previous method that attempted to solve the same problem as us. Note that the Scan2CAD method from \cite{avetisyan2019scan2cad}~(not to be confused with the Scan2CAD annotations) does retrieve CAD models for ScanNet, but from a very small pool created for benchmarking the proposed method: The size of the pool is taken to be equal to the number of objects in the scene, \ie, in a range of 5 to 30 CAD models selected from the ShapeNet dataset. It is also designed to contain the ground truth models. \cite{avetisyan2019scan2cad} shows an experiment with more CAD models~(400), it is however only a qualitative result. By contrast, we search through the entire set of CAD models for the target object class. For example, we consider the 8437 CAD Tables models and the 6779 CAD Chairs models in ShapeNet.

To the best of our knowledge, our method is thus the first one to achieve ground-truth-like CAD model retrieval and alignment. Overall, we make the following contributions:
\begin{enumerate}
\setlength\itemsep{0em}
    \item We introduce a fully-automatic method for CAD model retrieval and alignment that results in a fine alignment between the images and the reprojected CAD models.
    \item We show how to identify the objects that share the same geometry, and retrieve a common CAD model for these objects---we call this procedure 'cloning'. 
    \item We show that the quality of our results is comparable or better than human annotations.
    \item As no supervision is needed, our method can be directly used on various datasets for indoor scene understanding, and we show this by using two popular datasets, namely ScanNet~\cite{dai2017scannet} and ARKitScenes~\cite{dehghan2021arkitscenes}.
    \item Our method can be seen as a tool to automatically annotate images with CAD models and poses for the objects in the images, and we will make our annotations for the ScanNet scenes and for ARKitScenes publicly available. 
\end{enumerate}
\begin{table}[]
\centering
\scalebox{0.88}{
\begin{tabular}{@{}cccc@{}}
\toprule
Dataset      & \#scans & \begin{tabular}[c]{@{}c@{}} 3D GT\\ Labels\end{tabular} & \begin{tabular}[c]{@{}c@{}} CAD Model\\ Alignment\end{tabular} \\
\midrule
SceneNN~\cite{hua2016scenenn}     & 100  & OBB, Inst.Seg. & \xmark  \\
ScanNet~\cite{dai2017scannet}     & 1513  & Inst.Seg. & Scan2CAD~\cite{avetisyan2019scan2cad}~(*)  \\
Matterport3D~\cite{Matterport3D} & 2056  & Inst.Seg.  & \xmark  \\
ARKitScenes~\cite{dehghan2021arkitscenes}  & 5047   & OBB & \xmark \\ 
\bottomrule
\end{tabular}
}
\caption{\textbf{Overview of the most popular real RGB-D datasets for indoor scene understanding.}  'OBB' and 'Inst.Seg.' indicate that 3D oriented bounding boxes or 3D object instance segmentation are available as ground truth, respectively. None of the datasets directly provide full ground truth with CAD models, except for ScanNet thanks to the Scan2CAD dataset.  (*) Note however that the Scan2CAD dataset only provides CAD models for about 2 thirds of the objects.}
\label{tab:datasets_overview}
\end{table}

%-------------------------------------------------------------------------

\section{Related Work}

We focus our discussion on methods for CAD model retrieval in 3D scans. 

\subsection{Using Object Representations} 

A popular approach for CAD model retrieval is to represent the objects using embeddings. More exactly, it is possible to learn a representation for point clouds of objects, and a representation for CAD models in the same embedding space~\cite{avetisyan2019scan2cad, avetisyan2019end,dahnert2019embedding,Grabner20183DPE,hampali2021monte}. 

This approach allows for fast retrieval: The embedding space can be learned so that it can be used by efficient methods such as hashing for nearest neighbor search. However, annotated data is required to learn the embeddings. In our case, we are interested in an offline method;  computation time is thus not critical while accuracy is much more important. We thus consider an exhaustive search and a geometric criterion to evaluate how much a CAD model corresponds to the 3D scan. This is a safer approach than relying on the similarity between implicit embeddings.

\subsection{[Semi-]Manual Retrieval} 

A few datasets already exist for evaluating deep learning approaches for CAD model retrieval. 

The authors of Pix3D~\cite{sun2018pix3d} crawled web images to find images of objects from the IKEA object catalogue that were later verified by Amazon Mechanical Turk workers.  To simplify the 3D annotation procedure, they also kept only images without occlusions.  Then, human annotators would manually select keypoint correspondences between each input image and the 3D model manually selected for the image.  Using PnP and refinement, it is then possible to determine the 6D object pose.  Similarly, for creating the Scan2CAD dataset~\cite{avetisyan2019scan2cad}, human annotators were asked for each object to first, find the corresponding model in the ShapeNet dataset~\cite{chang2015shapenet} for the object within a bounding box and, then, click on correspondences between the 3D model and the point cloud to obtain the alignment.
Manually annotating 3D scans is however a heavy burden: Not all objects are annotated in Scan2CAD and ARKitScenes is not annotated with CAD models yet. 
We thus believe that our approach is very useful and scales better to large datasets.

Using synthetic images for training is also an attractive option, as the 3D annotations come then free. However, creating the 3D content and rendering it has also a high cost. For example, \cite{applepaper} reports a cost of \$57K to create Hypersim, a dataset of about 71’000 synthetic images for scene creation + image rendering, and a period of 2.4 years of wall-clock time on a large compute node.

\subsection{Analysis-by-Synthesis Approaches}
Analysis-by-synthesis in not a new concept in computer vision~\cite{yuille_abs,isola2013scene,hejrati2014analysis,huang2018holistic}. Some methods combine analysis-by-synthesis with learning~\cite{Kundu18, li-eccv18-deepim} but others, like us, do not~\cite{gupta2010estimating,ZouGLH19,hampali2021monte}. In particular, the recent \cite{hampali2021monte} is probably the closest work to us. However, \cite{hampali2021monte} focuses on retrieving arrangements of objects that explain the point clouds overall and not on accuracy. In our case, we exploit the existing 3D bounding box annotations as our goal is to provide accurate annotations. We perform an exhaustive search on the CAD models while \cite{hampali2021monte} relies on embeddings. We also provide the annotations we retrieved.
%-------------------------------------------------------------------------

\textvars{dpt,cad,msh,sns}
\textvars{Sil,CD}
\textvars{dist,ret,cham,thres}

\section{Method}

For each scene, we have an RGB-D scan of $N$ registered RGB images and depth maps, as well as a 3D mesh reconstructed from the scan. Our aim is to find for each object in the scene the most similar CAD model within the ShapeNet dataset, along with the corresponding pose as a 9-DoF transformation $T_i$, made of a 3D translation, 3D rotation, and scale along the 3 axes.
We only consider target objects that have either object box annotations \textbf{or} semantic instance segmentation available as supervision. Note that this kind of supervision is relatively easy to generate~(and often already available as ground truth in public 3D indoor scene datasets, as seen in Table~\ref{tab:datasets_overview}) compared to CAD model retrieval annotation, which requires the annotator to manually search a suitable CAD model from a large database. 

An overview of our method is shown in Figure~\ref{fig:overview}. It first performs CAD model retrieval, by sequentially replacing the target object in the scene with all available CAD models from the same object category. After replacing the object, we render the modified scene to generate observations corresponding to the selected CAD model. By doing this for all CAD models, we exhaustively search for the most perceptually similar model, according to the observations. After finding initial CAD models for each target object, our method focuses on retrieving geometrically similar CAD models for geometrically similar target objects in the scene. Finally, we perform differentiable object alignment to optimize the 9-DoF pose of each CAD model, taking into account the top-k retrievals for alignment-aware object retrieval.
We describe each component of our CAD model retrieval pipeline in detail below.

\subsection{Data Preprocessing}
\label{sec:preproc}

We use the 3D oriented object box as initialization for position, scale and orientation of the target object, whereas 3D semantic instance segmentation is used as geometrical information to replace the target object in the scene with the CAD models. We use either the 3D bounding box information or the instance segmentation provided by the dataset, which allows us to calculate an approximation for the other missing part.
We select a number $N_T$ of frames from all frames of the RGB-D scan, by selecting the frames where the target object is in the field of view according to its 3D bounding box and then regularly sample these frames.
We also use the provided class label from each target object to identify the corresponding ShapeNet category.

\subsection{Initial CAD Model Retrieval}
\label{sec:cad_model_retrieval}

\begin{figure}
      \centering
 		\includegraphics[trim={0cm 0.cm 0cm 0.cm},width=1.\linewidth]{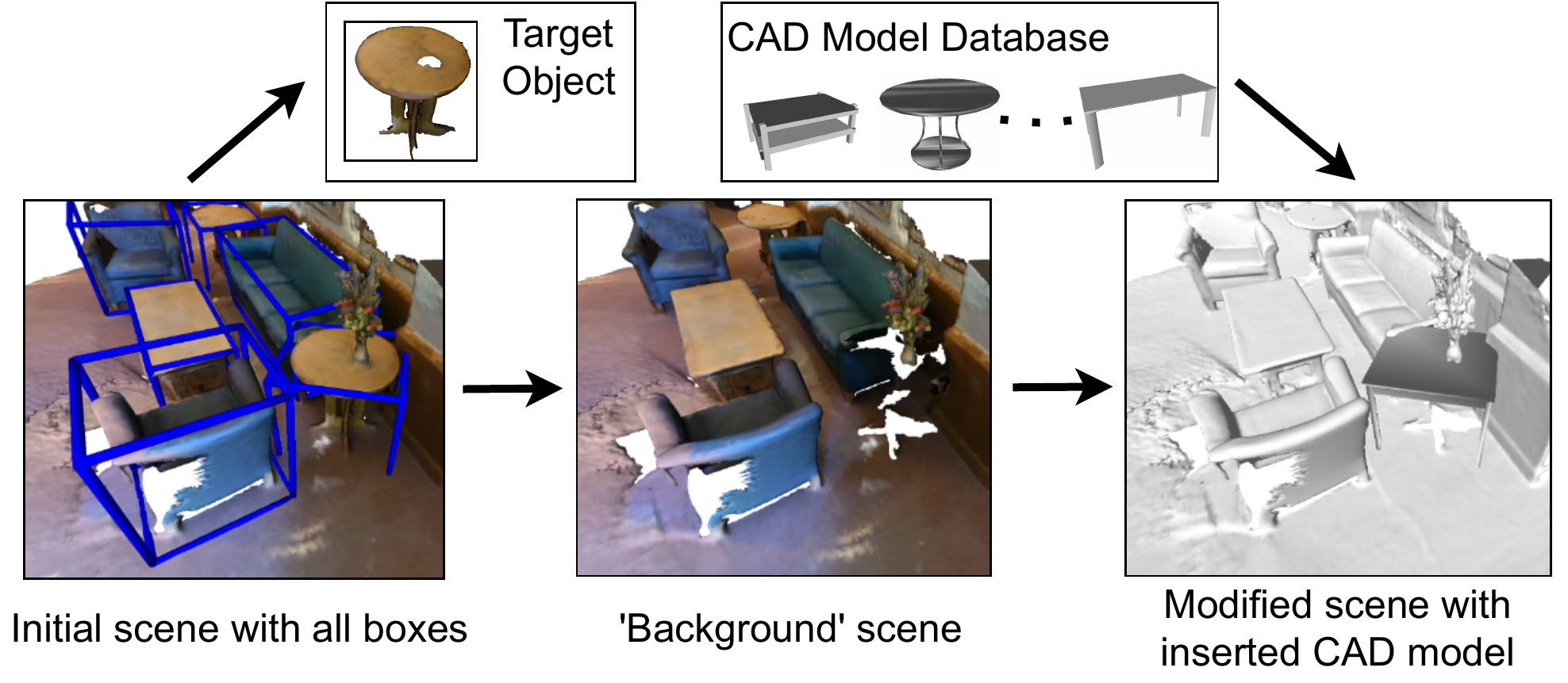}
	\caption[]{\textbf{Object replacement by removing the 3D points belonging to the real object  and inserting one possible CAD model.} The modified scene is then used to generate observations used for our objective function. }
	\label{fig:cut_mesh}
\end{figure}

We sequentially replace the target object in the input scene with each ShapeNet CAD model from the corresponding class category. Figure~\ref{fig:cut_mesh} illustrates this replacement procedure: We use the 3D semantic instance segmentation to remove all 3D points in the scene which are part of the target object $o_i$. Then,  we use the initial 3D scale, position and orientation extracted from the 3D oriented object bounding box to calculate an initial transformation which enables us to directly insert the CAD model $m_i$ at the position of the target object $o_i$. 

Our method follows an analysis-by-synthesis approach to evaluate how well a CAD model corresponds to a real object.  Such an approach is important to obtain a fine alignment of the reprojected CAD models with the input images. To measure quantitatively how well a CAD model fits a real object, we introduce a simple objective function that mainly compares observed depth data and the depth of the CAD model. This objective function does not require any training, and thus does not require any annotation.

\subsubsection{Depth Matching Term $\calL_\dpt$} 
\label{sec:depth_term}

The main terms of our objective function compare the depth maps for the CAD model and the 3D scene mesh after replacement with the observed depth maps. 

\begin{figure}
\centering
\scalebox{0.95}{
\begin{tabular}{ccc}
  \includegraphics[width=0.3\linewidth]{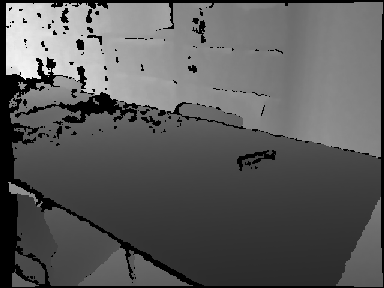} &
  \includegraphics[width=0.3\linewidth]{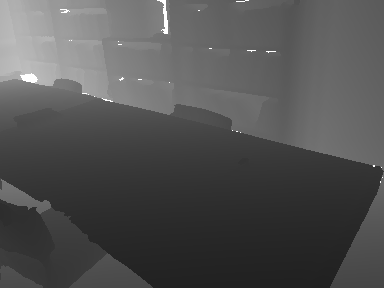} &
  \includegraphics[width=0.3\linewidth]{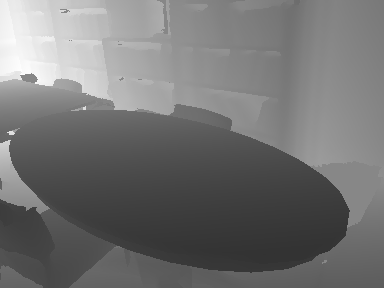} \\
Sensor depth $D_\sns$ &  Depth map $D_\msh$ & Depth map $D_\cad$ \\
 & from scene mesh & from scene mesh\\
 &                  & and CAD model\\
 &                  & after replacement\\
 &
 \includegraphics[width=0.3\linewidth]{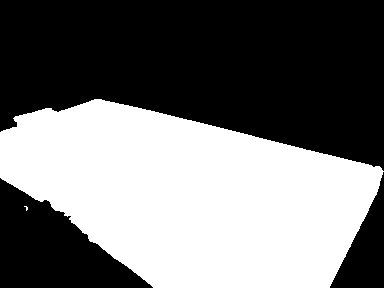} &
 \includegraphics[width=0.3\linewidth]{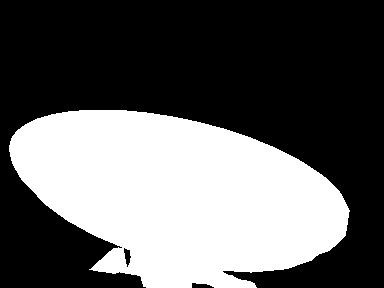} \\
 & Mask $S_\msh$ & Mask $S_\cad$ \\
 & from scene mesh & from CAD model\\
\end{tabular}
}
\caption{\textbf{Examples of the different depth maps (Top) and masks (Bottom)} used in our objective function. See Sections~\ref{sec:depth_term} and \ref{sec:add_terms}.}
\label{fig:2d_renderings}
\end{figure}

We noticed that with ScanNet, it is possible to generate a depth map of the scene that is better than the captured depth map by rendering the mesh, as seen in Figure~\ref{fig:2d_renderings}. For ARKitScenes, the scene meshes currently available are pretty noisy, and it is often better to consider the captured depth map than the rendered mesh. We therefore use a linear combination of the two possible comparisons:
\begin{equation}
  \label{eq:Ldpt}
\begin{aligned}
  \calL_\dpt = 
  \frac{1}{N_{T}} \sum_t \Bigl( &\frac{\lambda_m}{V_m^t} |M_m^t\cdot(D_\cad^t - D_\msh^t)|_1 \;+ \\
& \frac{\lambda_s}{V_s^t} |M_s^t\cdot(D_\cad^t - D_\sns^t) |_1 \Bigr) \> ,
\end{aligned}
\end{equation}
where the sum is over the $N_{T}$ selected frames of the RGB-D scan for the specific target object. Depth maps $D_\cad^t$, $D_\msh^t$, and $D_\sns^t$ are respectively the depth map rendered from the CAD model and the 3D scene mesh after replacement, the depth map rendered from the scene mesh, and the captured depth map for frame $t$.
$M_m^t$ and $M_s^t$ denote the valid pixel maps for the depth maps $D_\msh^t$ and $D_\sns^t$, respectively. We use the L1 norm to compare the depth maps 
and normalize the norms by the numbers of valid pixels $V_m^t$ and $V_s^t$. Weights $\lambda_m$ and $\lambda_s$ can be adapted to the dataset, depending on the quality of the captured depth maps and the 3D mesh.

Examples for $D_\cad^t$, $D_\msh^t$, and $D_\sns^t$ are shown in Figure~\ref{fig:2d_renderings}. 
Rendering a depth map $D_\msh^t$ is computed once per target object, whereas $D_\cad^t$ has to be computed for each CAD model that comes into consideration. However, this can be done efficiently by rendering the 'background' scene once, and then rendering each CAD model independently and fusing the results.

\subsubsection{Additional Terms $\calL_\Sil$ and $\calL_\CD$}
\label{sec:add_terms}
During our experiments, we noticed that in some specific cases, the term $\calL_\dpt$ introduced above is not sufficient to obtain optimal CAD model retrieval. For example, the term $\calL_\CD$ introduced below proved to be decisive when the selection of frames does not cover the whole target object, whereas the term $\calL_\Sil$ provide strong information about details of the object's shape. 

For $\calL_\Sil$, we consider the Intersection-over-Union between the silhouettes of the real object and of the CAD model:
\begin{equation}
\calL_\Sil = \frac{1}{N_{T}} \sum_t (1-\text{IoU}(S_\msh^t,  S_\cad^t)) \> ,
\end{equation}
where $S_\msh^t$ and $S_\cad^t$ are the rendered object masks for the real object and a possible CAD model, respectively, for frame $t$. Examples of $S_\msh^t$ and $S_\cad^t$ are shown in Figure~\ref{fig:2d_renderings}.

Additionally, we use the (one-way) Chamfer distance from the points on the real object to the CAD model:
\begin{equation}
\calL_\CD = \frac{1}{|P|} \sum_{p \in P} \min_{q \in Q}  \|p - q\| \> , 
\end{equation}
where $P$ is the point cloud for the real object as identified using the bounding box and/or the segmentation as explained in Section~\ref{sec:preproc} and $Q$ is a set of 3D points sampled on the CAD model.

\subsubsection{Objective Function}
Our objective function can now be defined as
\begin{equation}
\calL = \calL_\dpt + \lambda_\Sil \calL_\Sil + \lambda_\CD \calL_\CD \> ,
\label{eq:objfun}
\end{equation}
where $\lambda_\Sil$ and $\lambda_\CD$ are weights. We compute this objective function for each CAD model in ShapeNet that matches the class label of the current target object. 

\subsection{CAD Model Cloning}
\begin{figure}[] 
 %trim={<left> <lower> <right> <upper>}
      \centering
 		\includegraphics[trim={0cm 0.cm 0cm 0.cm},width=.9\linewidth]{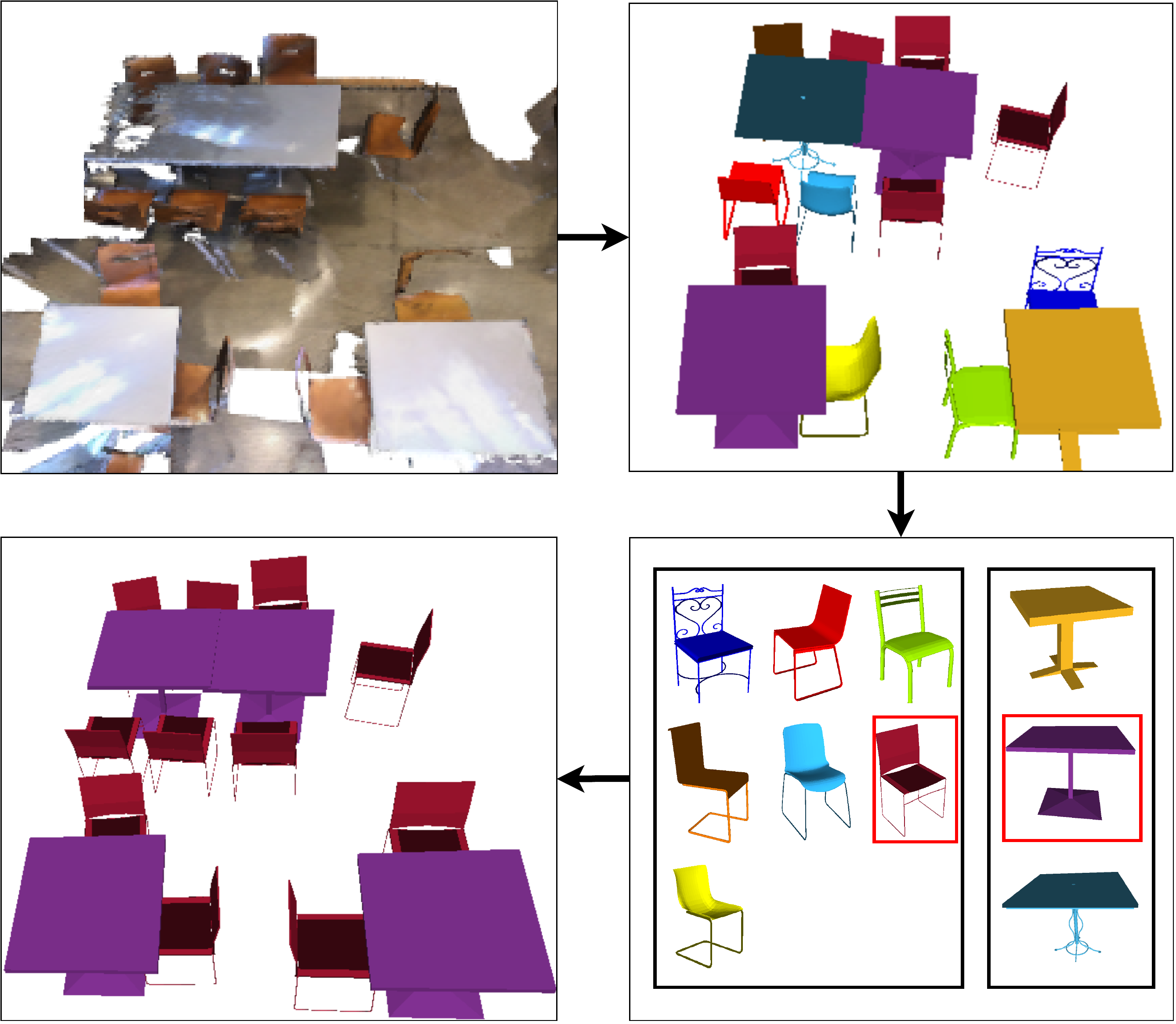}
	\caption[]{\textbf{CAD model cloning.} We first look for a CAD model for each object independently. We then cluster the CAD models based on their shape similarity. Clusters with more than one CAD model indicate objects that share the same shape. We then assign to all the objects in the cluster the CAD model that minimizes the sum of the objective functions over all the objects.}
	\label{fig:CAD_sub}
\end{figure}

Given a 3D scene such as the one shown in Figure~\ref{fig:CAD_sub} for ScanNet, a human annotator will recognize that several objects in the scene share the same geometry. However, this behaviour is not guaranteed with our per-object predictions for CAD model retrieval. To solve this issue, we add a simple yet effective method to identify objects that share their shapes and to retrieve an appropriate CAD model for all the identified objects. 

We first cluster the CAD models retrieved independently for each object by  the method described in Section~\ref{sec:cad_model_retrieval}. We use a bottom-up clustering based on the pair-wise Chamfer distance between the CAD models. More precisely, for each pair of retrieved CAD models, we first calculate their symmetric Chamfer distance. Then, we iteratively go through all object pairs from lowest to highest Chamfer distance and perform one of the following steps if their Chamfer distance is below a threshold $\tau$~(we use $\tau = 3.10^{-3}$ in practice):
(a) If none of the two objects is already part of a cluster, a new cluster with these two objects is created.
(b) If one of the objects is already part of a cluster, and the other one is not assigned to a cluster yet, it is added to the same cluster.
(c) If both objects are already assigned to different clusters, these clusters are merged.

Then, for each cluster with more than one object, we look for a common CAD model that minimizes the sum of the objective functions $\calL$ from Eq.~\eqref{eq:objfun} over all the objects in the cluster, when replacing every object by the CAD model. Figure~\ref{fig:CAD_sub} shows an example for our CAD model clustering and cloning.

\subsection{Differentiable Pose Refinement}
Once we found a CAD model for an object, possibly by the cloning procedure explained above, we refine its 9-DoF pose $T$, \ie, its scale, position, and orientation to fit the depth observations better. We initialize its pose using the 3D bounding box that is either part of the dataset as ground truth, or calculated during our preprocessing step. 
To do this, we minimize the objective function of Eq.~\eqref{eq:objfun} over the 9 pose parameters in $T$ using the 
differentiable rendering pipeline of \cite{ravi2020pytorch3d} and the Adam optimizer~\cite{kingma2014adam}.

\section{Evaluation and Experiments}

To evaluate the performance of our method, we considered the popular ScanNet dataset and the more recent ARKitScenes dataset. We asked computer vision experts to conduct a visual evaluation on ScanNet to compare our automatically retrieved annotations with the annotations provided by Scan2CAD, to evaluate if our method can provide ground truth annotations with comparable quality to manual annotations. We then conduct a visual inspection of the results on ScanNet, and additionally provide qualitative results on the ARKitScenes dataset. 

\subsection{Evaluation using Scan2CAD}

To evaluate the performance of our proposed method, we conducted a visual evaluation by asking 3D experts to compare our results to Scan2CAD, which provides ground truth for CAD model retrieval and alignment for ScanNet. We perform this evaluation on the ScanNet validation set, which consists of $312$ indoor scenes. To ensure fair comparison, we use the initial 3D bounding boxes from Scan2CAD. 

\paragraph{Implementation details.} We set the weights in Eqs.~\eqref{eq:Ldpt} and \eqref{eq:objfun} to
$\lambda_\msh=0.75, \lambda_\sns = 0.9 $, $\lambda_\Sil=0.3$ and $\lambda_\CD=2.0$. To compute the Chamfer distance in $\calL_\CD$, we uniformly sample $N=10k$ points from the surface mesh of each CAD model.  
For the CAD models that were not substituted by our cloning method, we first retrieve the top-3 best CAD models, perform pose refinement for each 3 of them, and keep the one with the lowest objective function value.
For the CAD models issued by the cloning procedure, 
we perform the pose refinement with the common CAD model.

\subsubsection{Quantitative Comparison}
\label{sec:quant}

As proposed in the Scan2CAD paper~\cite{avetisyan2019scan2cad}, CAD retrieval and alignment can be evaluated in terms of translation, rotation, and scale error. Additionally, we argue that shape difference is also a important criterion. For each target object in the validation set, we compare our results with the corresponding Scan2CAD ground truth, by calculating the difference of rotation, translation, scale and shape.

\begin{figure}[] 
 %trim={<left> <lower> <right> <upper>}
      \centering
 		\includegraphics[trim={0cm 0.cm 0cm 0.cm},width=1.\linewidth]{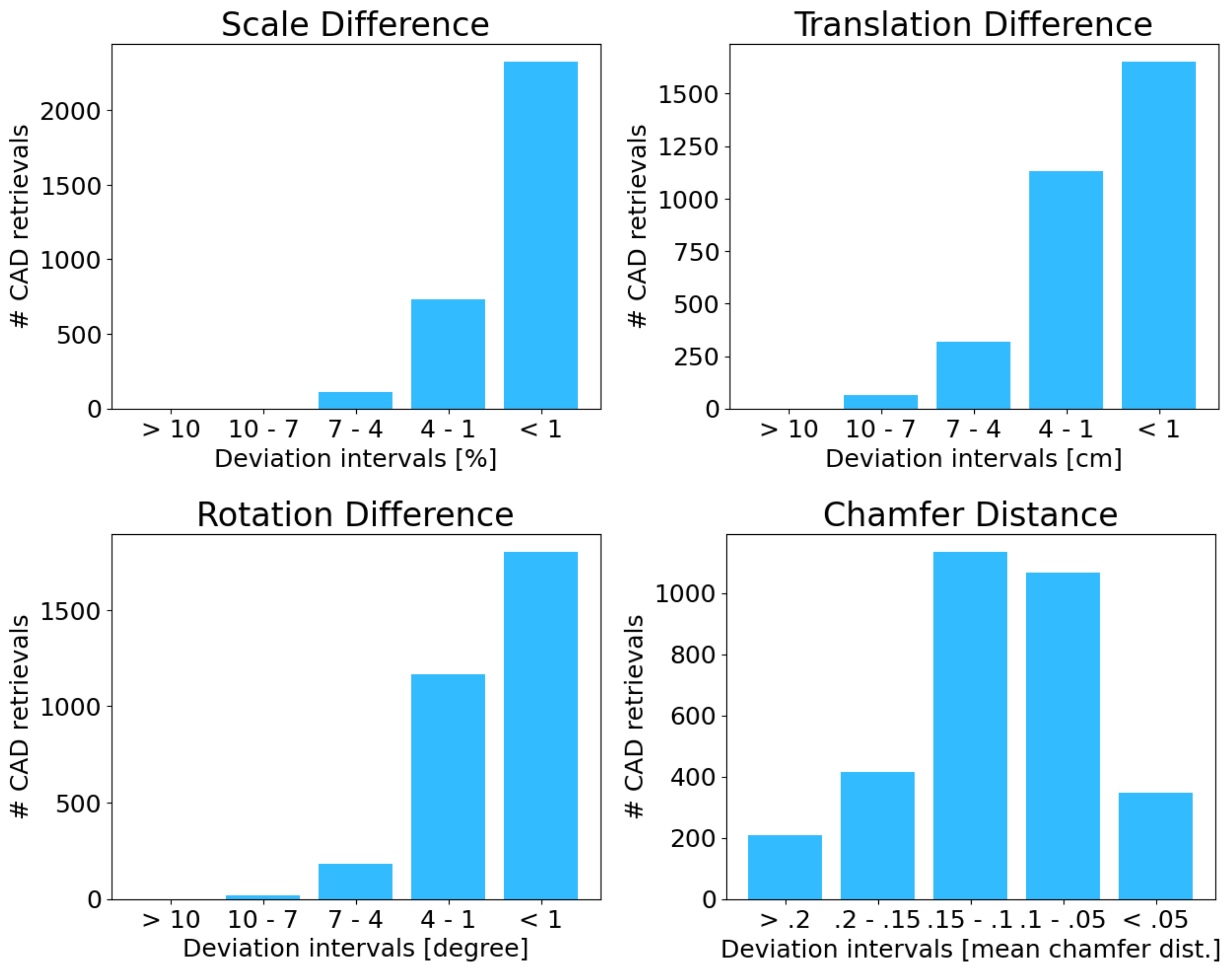}
	\caption[]{Deviations of our results compared to Scan2CAD ground truth for each criterion on the entire ScanNet evaluation dataset. The majority of CAD retrievals are very close to their Scan2CAD counterpart, indicating the high quality of our results. }
	\label{fig:error_plot}
\end{figure}

Figure~\ref{fig:error_plot} shows the deviation of our results compared to the Scan2CAD annotations. As one can see, the majority of our predicted CAD model retrievals do have a very small difference compared to ground truth for all criteria, which already indicates the high quality of our results. 

\subsubsection{Visual Evaluation by Experts}
\label{sec:vis_eval_experts}
\begin{figure}
 %trim={<left> <lower> <right> <upper>}
      \centering
  		\includegraphics[width=\linewidth]{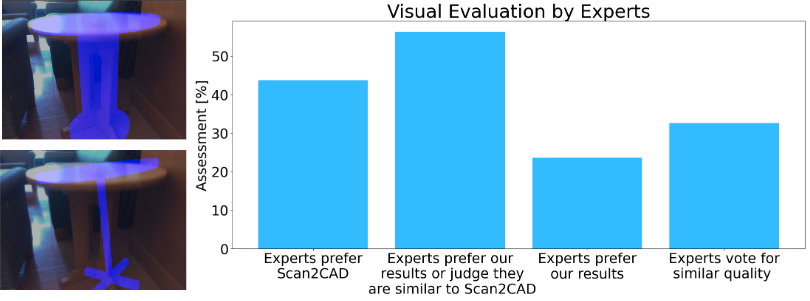}
	\caption{\textbf{Left: One example from our visual evaluation of the 165 objects with the largest deviations from the Scan2CAD annotations. }  The experts were shown two images, created by reprojecting CAD models in the same view from the RGB-D scan. We use the CAD models and their poses as provided by Scan2CAD for one image, and the models and poses retrieved by our method for the other image. The experts were not told which method was used and images were shuffled so that our method appeared 50\% of the time in the top image and 50\% of the time in the bottom image. For each pair of images, the experts were asked to answer the question: ``Do the reprojections in the top image look better, worse, or similar to the reprojection in the bottom image?''  \textbf{Right: Histograms of the answers.} On average for these 165 objects with the largest deviation, our method provides similar or better CAD models, while being automatic.
	}
	\label{fig:study}
\end{figure}

The quantitative comparison with the Scan2CAD annotations provided above gives the deviation with respect to these annotations, however it does not tell what these differences mean in terms of overall quality, and which annotations are better---or if they are equivalent.

To answer these questions, we conducted a visual evaluation by computer vision experts on the 165 objects with the largest deviation from the Scan2CAD annotations. To keep the comparison simple, we generated pairs of images where we reprojected the CAD models provided by Scan2CAD and the CAD models retrieved by our method. Figure~\ref{fig:study} shows an example of such a pair of images. Such reprojections were easier to interpret for the subjects than a 3D scene, but to make sure that the poses of the CAD models were also evaluated correctly and not just their reprojections, we used multiple views of the same scenes.

The experts were not told which method was used and images were shuffled so that our method appeared 50\% of the time in the top image and 50\% of the time in the bottom image, to prevent the subjects to be biased towards one of the two sides or methods. For each pair of images, the experts were asked to answer the question: ``Do the reprojections in the top image look better, worse, or similar to the reprojection in the bottom image?''  As this question requires some domain expertise, we had to restrict the subjects to PhD students working on computer vision and computer graphics. In total, 9 experts participated to this study.

The results are shown in Figure~\ref{fig:study}: For 43.7\% of all the shown pairs, the subjects prefer the CAD models and poses from Scan2CAD, for 24.5\%, they prefer the CAD models and poses retrieved by our method, and for the 31.8\% left, they found the two sources were similar.  This validates that our method can be used in place of human annotations, as it produces annotations of similar or better quality on average.

\subsubsection{Fine-grain Evaluation}

\begin{figure}
 %trim={<left> <lower> <right> <upper>}
      \centering
 		\includegraphics[trim={0cm 0.cm 0cm 0.cm},width=.9\linewidth]{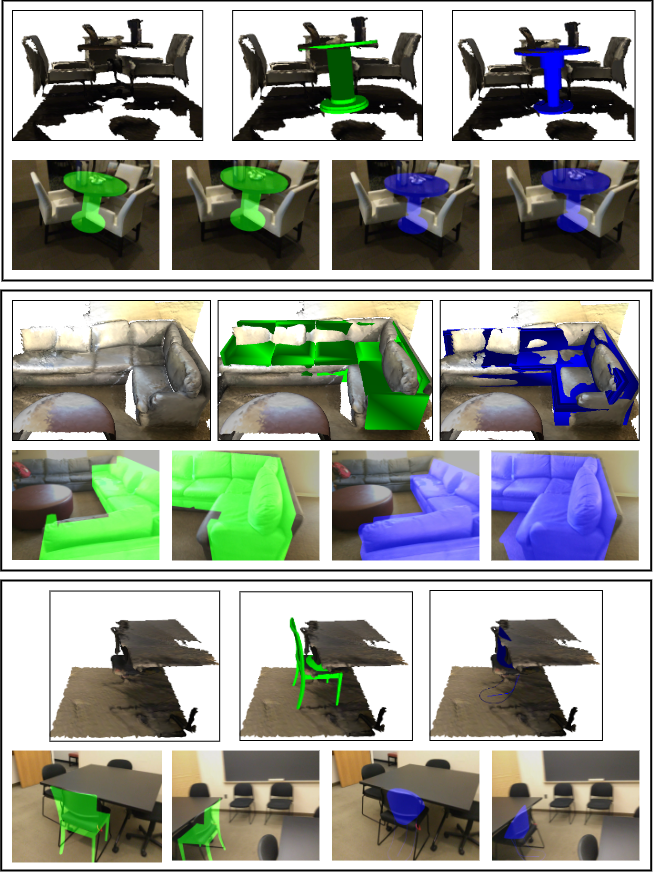}
	\caption[]{
\textbf{Visualization of some  objects} with the largest difference between the Scan2CAD annotations and the annotations created by our method.
We show the RGB-D scan, the 3D overlay of the Scan2CAD object (in green) and the 3D overlay of the CAD model retrieved with our method~(in blue), and below, the 2d reprojections of the CAD models. First example: Rotation difference of $6.33^{\circ}$, Assessment: our method is more accurate. Second example: Translation difference of $8.5cm$, Assessment: equal quality. Third example: Chamfer distance of $0.42$, Assessment: Scan2CAD annotation is more accurate.}
		\label{fig:qualitative_results}
\end{figure}

To get a better understanding of how our automatic annotations compare to the manual ones,
we visualized the $50$ objects with the highest deviation for each criterion considered in Section~\ref{sec:quant}~(translation, rotation, scale, shape) and evaluated the geometry and alignment through visual inspection. Table~\ref{tab:eval_vis_insp} shows the results. For the majority of these objects, the overall quality of geometric shape and alignment is equal or better than the annotations from the Scan2CAD dataset. Figure~\ref{fig:qualitative_results} shows examples for objects with large differences for different criteria.

\begin{table}
\centering
\scalebox{0.75}{
\begin{tabular}{@{}ccccc@{}}
\toprule
          &             &               &              & Annotations\\ 
          & Scan2CAD    & Annotations   & Annotations  & by our method\\
          & annotations & by our method & have similar & are similar\\
Criterion & are better  & are better    & quality      & or better\\ 
\midrule
Translation & 10        & 7  & 33 & 40 \\ 
Rotation        & 12        & 10  & 28 & 38 \\ 
Scale & 12        & 7  & 31 & 38 \\ 
Shape & 20        & 8  & 22 & 30 \\ 
\bottomrule
\end{tabular}
}
\caption{\textbf{Results of the visual inspection} of the 50 objects with the largest deviation from the Scan2CAD annotations. A clear majority of annotations are at least equal in quality.}
\label{tab:eval_vis_insp}
\end{table}

\subsubsection{Additional Annotations for ScanNet}
\label{sec:scan2cad_enhance}

\begin{figure} 
 %trim={<left> <lower> <right> <upper>}
      \centering
 		\includegraphics[trim={0cm 0.cm 0cm 0.cm},width=.88\linewidth]{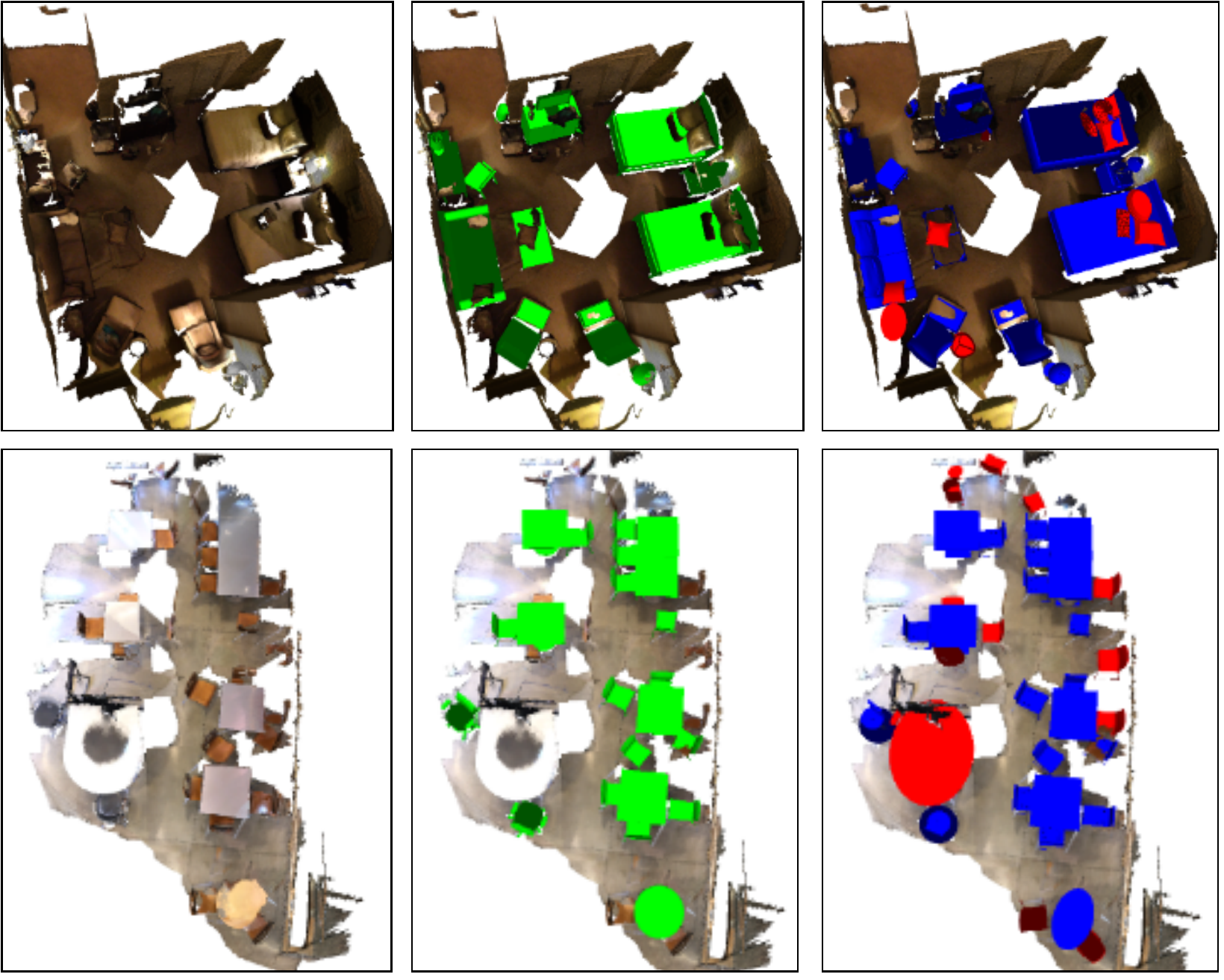}
	\caption[]{\textbf{Comparison with Scan2CAD annotations.} Left: RGB-D scan. Middle: Scan2CAD annotations. Right: Our results, where red CAD models are for objects not annotated in Scan2CAD.}
	\label{fig:CAD_Scan2CAD_enhance}
\end{figure}

Figure~\ref{fig:CAD_Scan2CAD_enhance} shows a comparison of results for full scene CAD retrieval by our method with Scan2CAD: As already mentioned, the Scan2CAD dataset does not provide annotations for all the objects.  About one third of the objects of standard classes~($1130$ objects for the $312$ scenes in the ScanNet validation set) are not labeled with a CAD model nor pose, but since our method is automated, we were able to retrieve a CAD model and pose for all these objects.  This represents on average $3.6$ additional objects per scene, and about 37\% more retrieved objects than in the Scan2CAD annotations.

\subsection{Results on the ARKitScenes Dataset}

\begin{figure} 
 %trim={<left> <lower> <right> <upper>}
      \centering
 		\includegraphics[trim={0cm 0.cm 0cm 0.cm},width=.88\linewidth]{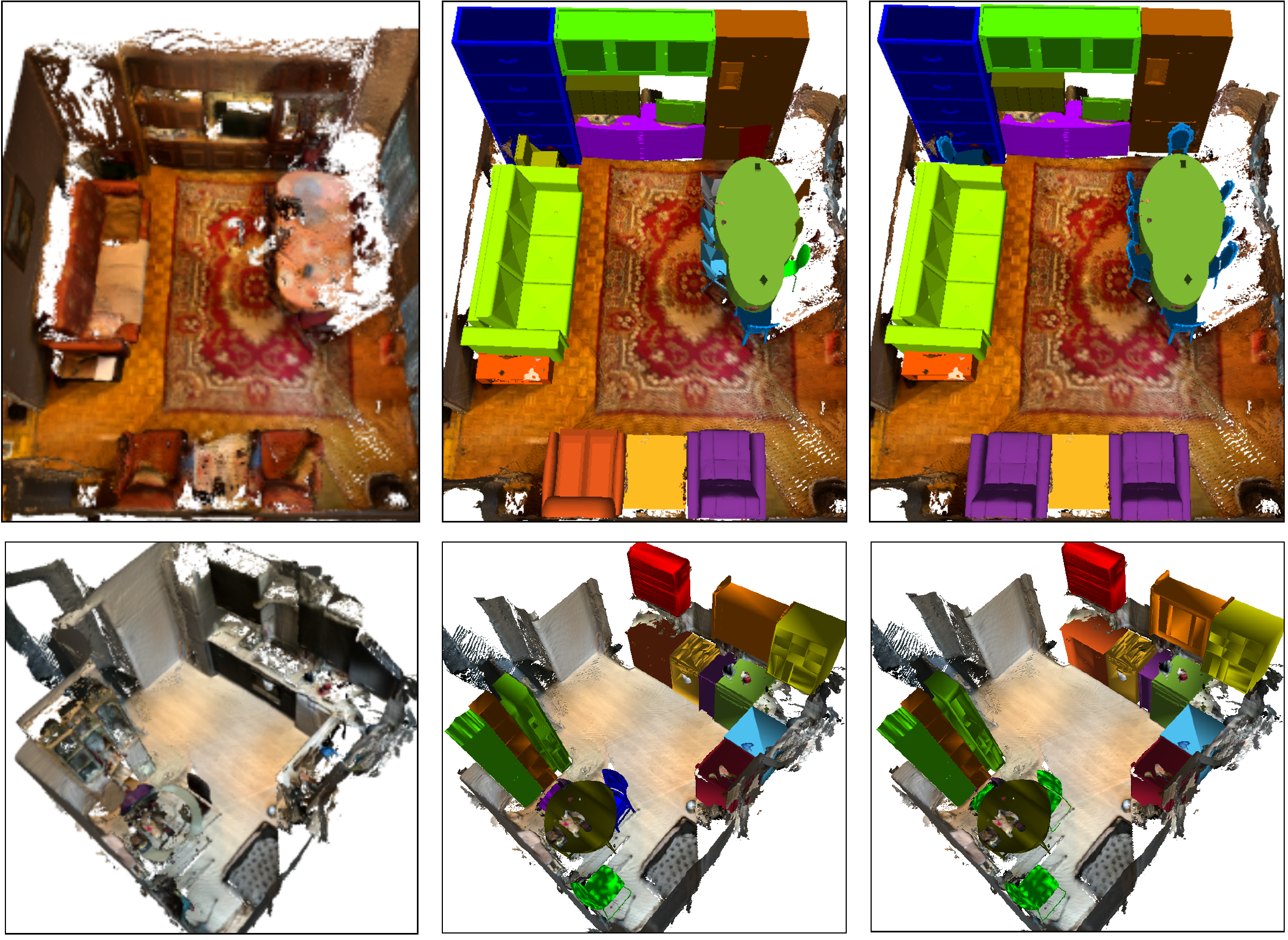}
	\caption[]{\textbf{Qualitative results on the ARKitScenes dataset.}  Left: RGB-D scan. Middle: Our results before cloning. Right: After cloning. CAD models with the same color have the same geometry: In the top row, our cloning procedure correctly retrieves the same models for the chairs and the two armchairs at the bottom.}
	\label{fig:results_ARKit_scenes}
\end{figure}

To show the generalization ability of our method, we also ran our method on the recently published ARKitScenes dataset. This dataset consists of $5047$ RGB-D scans of indoor scenes, and provides 3D oriented bounding boxes as
ground-truth annotations. Note that there are no CAD models provided as ground truth for this dataset.
\paragraph{Implementation details.} The quality of the sensor data for this dataset is significantly better compared to ScanNet, 
and the depth maps provided by ARKitScenes are much better. However, the quality of the 3D scene meshes provided by ARKitScenes is still significantly lower than the ones available for ScanNet.  Therefore, we adjusted the weight parameters to
$\lambda_s = 1.3$, $\lambda_m = 0.3$, $\lambda_\Sil = 0.4$, $\lambda_\CD=1.5$.
\paragraph{Retrieved annotations.}~Figure~\ref{fig:results_ARKit_scenes} shows examples of the CAD models and their poses retrieved by our method. The advantage of our cloning procedure is also clearly visible.

%-------------------------------------------------------------------------

 \section{Conclusion}
We presented a method to retrieve CAD models for objects in 3D scans and their poses that have similar quality as manual annotations. 
The CAD models reproject well on the images used to capture the scans. We thus hope that our results can be used for 3D scene understanding from single images. We will make available the annotations we retrieved for the ScanNet and ARKitScenes datasets.\\
\textbf{Acknowledgment.} This work was supported by the Christian Doppler Laboratory for Semantic 3D Computer Vision, funded in part by Qualcomm Inc.

%-------------------------------------------------------------------------

{\small
\bibliographystyle{ieee_fullname}
\bibliography{0_cad_retrieve}
}

\end{document}